\documentclass{sig-alternate-05-2015}
\usepackage{balance}
\usepackage{graphicx}
\usepackage{amssymb}
 \usepackage{epsfig}
\usepackage{amsmath}
\usepackage{flushend}
\usepackage{fancybox}
\usepackage{subfig}
\usepackage{url}
\usepackage[usenames,dvipsnames]{pstricks}
\usepackage{pstricks-add}
\usepackage{pstricks,pst-node,pst-tree}
\usepackage{multirow} 
\usepackage{wrapfig}
\usepackage{subfig}
\newcommand{\alg}{{\sc PRIIME}}
\newcommand{\argmax}{\arg\!\max}
\newcommand{\argmin}{\arg\!\min}
\makeatletter
\def\hlinewd#1{%
\noalign{\ifnum0=`}\fi\hrule \@height #1 
\futurelet\reserved@a\@xhline}
\makeatother

\begin{document}
 
%
%

\title{PRIIME: A Generic Framework for Interactive Personalized Interesting Pattern Discovery}
\numberofauthors{2} 
%
\author{
\alignauthor Mansurul A Bhuiyan\\
       \affaddr{Dept. of Computer Science, Indiana University---Purdue University}\\
       \affaddr{Indianapolis, USA}\\
       \email{mbhuiyan@iupui.edu}
\alignauthor Mohammad Al Hasan\\
       \affaddr{Dept. of Computer Science, Indiana University---Purdue University}\\
       \affaddr{Indianapolis, USA}\\
       \email{alhasan@cs.iupui.edu}
}
\maketitle

\begin{abstract} 

The traditional frequent pattern mining algorithms generate an exponentially large number
of patterns of which a substantial proportion are not much significant for many
data analysis endeavors. Discovery of a small number of personalized interesting 
patterns from the large output set according to a particular user's
interest is an important as well as challenging task. Existing works on pattern 
summarization do not solve this problem from the personalization viewpoint. 
In this work, we propose an interactive pattern discovery framework named \alg\ 
which identifies a set of interesting patterns for
a specific user without requiring any prior input on the interestingness
measure of patterns from the user. The proposed framework is generic to support
 discovery of the interesting set, sequence and graph type patterns.
We develop a softmax classification based iterative learning algorithm that uses a limited number of
interactive feedback from the user to learn her interestingness profile, 
and use this profile for pattern recommendation. To handle sequence
and graph type patterns \alg\ adopts a neural net~(NN) based unsupervised feature 
construction approach. We also develop a strategy that combines exploration
and exploitation to select patterns for feedback. 
We show experimental results on several real-life datasets to validate the performance of the
proposed method. We also compare with the existing methods of interactive pattern
discovery to show that our method is substantially
superior in performance. To portray the applicability of the
framework, we present a case study from the real-estate domain.
\end{abstract}


\section{Introduction}\label{sec:introduction}
Frequent pattern (itemsets, sequences, or graphs)
mining~\cite{zaki2014dataminingbook} has been a core research task in data
mining domain for over two decades, yet its deployment in real-life data
analysis has been moderate due to the following two challenges: (1) the pattern
space is combinatorial, so the number of patterns that a mining task produces
is generally too large to process by an end-user; (2) for various data analysis
tasks, frequency threshold is not a sufficient filter criterion for selecting
patterns. To overcome (1), many works have been proposed for pattern
summarization and compression~\cite{Mampaey.Vreeken:2012,Vreeken.Leeuwen:2011},
and to overcome (2), several alternative {\em interestingness} metrics, such
as, Jaccard index, odds ratio, and lift~\cite{zaki2014dataminingbook} have been proposed which can be used
together with frequency. Nevertheless, interesting pattern discovery still
remains an unsolved problem  due to the subjectivity in the definition of
\textit{interestingness}.  In many cases, for a specific user, off-the-shelf
interestingness metrics does not represent her true interestingness over
patterns. Pattern summarization does not help either, as such an approach
solves interesting pattern discovery from a global perspective which is far
from personalization what is needed to meet the pattern discovery demand of a
specific user.


There exist a few works which target personalized pattern discovery by using
users
feedback~\cite{Xin.Shen.ea:06,boley.Mampaey:2013,Bhuiyan.Hasan:12}.
The overall methodologies of these works are to build an interactive pattern
discovery system, that works as the following---a user provides ratings on a
small collection of patterns, then the system uses these rating to train a
personalization model, which is later used to isolates user's preferred
patterns from the remaining patterns. Major design choices of an interactive
pattern discovery system are: (1) genericness, i.e. the kinds of patterns that
the system supports; (2) pattern selection process for feedback---whether it is
active~\cite{settles2010active} or non-active; (3) learning model that the system uses; and (4) feature
representation of patterns for facilitating learning.  Existing solutions for
interactive pattern discovery are poor on multiple among the above design
choices. We elaborate more on this in the following paragraphs.

The majority of the existing platforms for interactive pattern discovery only
support itemset patterns because they do not have an effective metric embedding
for more complex patterns, say, sequences, or graphs. Metric embedding of
patterns is needed for facilitating the training of a model which discriminates
between interesting and not-so-interesting patterns.  For itemset patterns,
existing works leverage bag of items for instrumenting a metric representation.
However, no such natural instrumentation is available for complex patterns,
such as a sequence, or a graph. Sometimes, n-grams are used for metric embedding of
sequences~\cite{aggarwal2014data}, and topological measures, such as
centralities, eccentricity, egonet degree, egonet size, and diameter are used
for feature representation of graphs~\cite{Li:2012}. While the above feature
representation may work well for the task of traditional sequence or graph
classification, they do not work well for frequent patterns which are numerous and
small in size. So, a generic yet principled approach is needed for metric
embedding of complex patterns, such as sequences and graphs.

For any interactive learning platform, designing a method for selecting a small
set of patterns for which a user's feedback is sought is critical. Given that the
number of frequent patterns is typically enormous, the performance of interactive learning
depends critically on the module that selects the patterns for collecting
user's feedback. Most of the existing interactive pattern 
discovery works~\cite{Xin.Shen.ea:06,Bhuiyan.Hasan:12}  
heavily entangle model learning with the feedback set selection; i.e. 
the influence of a model which is trained on the prior sets 
of feedback is high when they select patterns 
for a user's feedback. So, the initial sets of feedback play a significant role to 
the model's configuration hence, the model may risks suffering 
from the {\em positive reinforcement} phenomenon. In other words, 
high influence of the current model to select patterns for obtaining the user's feedback 
in the next iteration is similar to exploitation in 
active learning~\cite{Bondu.Lemaire.ea:10}. 
It sometimes makes the predictive model represents some areas of the pattern space, 
making it sub-optimal over the entire pattern space. 

In this work, we propose a generic framework for interactive personalized
interesting pattern discovery~(IPIPD) called \alg \footnote{\alg\ is an anagram of the bold letters in gene\textbf{R}ic 
fra\textbf{M}work for \textbf{I}nteractive p\textbf{E}rsonalized
\textbf{I}nteresting \textbf{P}attern discovery} by incorporating a user in
the discovery pipeline.  The proposed method uses preference rating of a small
set of frequent patterns from the user to learn her interestingness criteria and
recommend a set of patterns which incline the most towards her choice.  To
address the challenges discussed above, we develop a neural net based
unsupervised feature vector representation scheme for sequence and graph
patterns. We propose a pattern selection approach for feedback to combat
the {\em positive reinforcement} phenomenon.  We also assume that feedback is
given using positive real discrete values, where a higher value stands for more
empathy towards the given pattern.  For example, a user can have a three-class
rating system, where 1, 2 and 3 means dislike, not sure and likes the pattern. 

In recent years, unsupervised feature learning using traditional and deep
neural networks has become popular. These methods help an analyst discover
features automatically, thus obviating feature engineering using domain
knowledge. Researchers achieve excellent performance with these unsupervised
learning techniques for extraction of features from text~\cite{Mikolov:2013},
speech~\cite{socher:2011} and images~\cite{Szegedy:2014}. Recently
in~\cite{Perozzi:2014}, authors have shown the potential of building
unsupervised feature representation of the vertices of an input graph for
vertex classification~\cite{Perozzi:2014}. \alg's metric representation method
for graphs and sequences is motivated from these works. Specifically, for
a set of sequence or graph patterns, \alg\ maps each pattern into a sentence and
pattern elements (items in a sequence or edges in a graph) as words. Afterwards, 
it leverages a language model similar
to~\cite{le2014distributed} for finding $d$-dimensional feature vector
representation of the patterns. We experimentally show that such technique
performs better than the existing ad-hoc metric embedding.


\alg\ also provides a robust pattern selection strategy for rating
solicitation. Unlike earlier interactive pattern discovery
works~\cite{Xin.Shen.ea:06,Bhuiyan.Hasan:12}, for the case of \alg, the coupling
between the existing state of the learning model and the training set selection
is significantly reduced which yields a much-improved learning model. 
In fact, \alg's strategy is a combination of exploration and
exploitation which enables it to reduce the bias that existing state of the
learning model imposes on pattern selection. Exploitation aspect of the
proposed strategy prefers patterns (for feedback solicitation) that are likely
to influence the learning model the most, i.e. have the largest impact on
model's parameters.  On the other hand, the exploration aspect ensures that a
broader span of the pattern space is covered.  \alg\ maximizes expected
gradient length~(EGL)~\cite{Settles.Craven:2008} (an active learning strategy)
for exploitation and it chooses a diverse set of patterns using $k$-center~\cite{Gonzalez:85} 
search for exploration.  We empirically show that a
combination of above two triggers better learning performance 
than the existing interactive pattern discovery frameworks.


We claim the following contributions:

\noindent $\bullet$ We propose a generic interactive personalized interesting pattern
discovery framework called \alg\ that is based on iterative learning of a user's
interestingness function using the supervised classification model. The
system does not require any prior knowledge of the user's interestingness
criteria and needs a limited degree of user engagement.

\noindent $\bullet$ We develop an efficient unsupervised feature construction
scheme for sequence and graph patterns and propose a pattern selection 
strategy combining exploitation and exploitation for feedback collection.

\noindent  $\bullet$ We perform exhaustive empirical validations of the
proposed framework using real-world set, sequence, and graph datasets. We show
that the proposed feature construction and pattern selection 
performs better than the existing ones. To show the applicability of the 
interactive solution of personalized interesting pattern discovery, 
we present a case study from the real-estate domain using a real-life housing data.

The reminder of this paper is organized as follows. In Section~\ref{sec:architecture},
we discuss in detail of the proposed framework. In Section~\ref{sec:method}, we talk
about the learning algorithm of the framework including pattern representation~(Subsection~\ref{sec:data-rep}),
classification model~(Subsection~\ref{sec:model}) and feedback mechanism~(Subsection~\ref{sec:representative}).
In Section~\ref{sec:experiment}, we present an extensive empirical evaluation including 
the performance of the learning algorithm~(Subsection~\ref{sec:performance}), 
neural net based unsupervised feature construction~(Subsection~\ref{sec:feature_cons}) and comparisons
with the existing methods~(Subsection~\ref{sec:compare}). In Section~\ref{sec:casestudy},
we present a case study to establish the real life application of the proposed
framework in a real-estate domain problem.
We conclude the paper with related works~(Section~\ref{sec:related}) 
and a conclusion~(Section~\ref{sec:conclusion}).

\begin{figure}[!ht]
\centering
\includegraphics[width=0.5\textwidth]{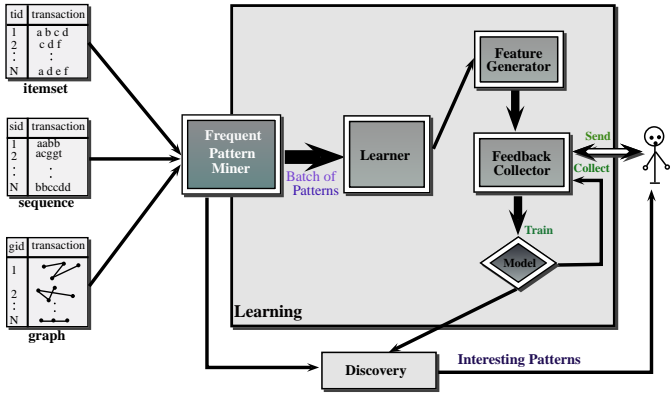}
\vspace{-0.15in}
\caption{Generic Interactive Personalized Interesting Pattern Discovery Framework}
\vspace{-0.18in}
\label{fig:framework}
\end{figure}


\section{Problem Definition and System Architecture }\label{sec:architecture}

Consider, a transactional dataset $\cal{D}$; each transaction in $\cal{D}$ is a
combinatorial object, such as an itemset, a sequence, or a graph. Depending on
the kind of objects in $\cal{D}$, it can be an itemset dataset, a sequence
dataset or a graph dataset.  A user $u$ is interested in mining interesting
patterns from $\cal{D}$, where the patterns (denoted by set ${\cal O}$) are
sub-objects, say subsets, subsequences, or subgraphs, over the objects in
$\cal{D}$. Traditional approaches consider frequency as the interestingness metric and
design frequent pattern mining algorithms which return patterns that exceed
minimum support thresholds over the transactions in $\cal{D}$. However, due to
the fact that the frequent pattern space is combinatorial, existing frequent
pattern mining algorithms generally return a large number of patterns
causing information overload.  Interactive pattern mining is a framework for
negotiating information overload so that a succinct set of interesting patterns
(a small subset of frequent patterns, ${\cal O}$) can be delivered to $u$,
which are personalized by utilizing $u$'s feedback on a small number of
patterns. We call this framework IPIPD, which stands for interactive
personalized interesting pattern discovery. \alg, our proposed method in this
paper follows this framework.

We assume, $u$ possesses an interestingness function (say, $f$) over the
frequent patterns, ${\cal O}$).  $f$ maps each pattern in ${\cal O}$ to a
positive integer number indicating interestingness criteria of the user $u$
i.e., $f: {\cal O} \rightarrow \mathbb{Z}_+$. \alg\ learns $f$ through an
active learning process over multiple learning iterations. During an iteration
(say $i$), the system returns a set of patterns $\{p_t\}_{1\le t \le l}$ to
$u$, which is selected from a partition of ${\cal O}$ (which constitutes the
$i$'th batch of pattern-set); $u$ sends feedback $\{y_t=f(p_t)\}_{1\le t \le
l}$ using an interactive console; $y_t$'s are positive integer numbers
reflecting the user's empathy towards the selected patterns; using the feedback
the system updates its current model of the user's interestingness function,
$f$.  An alternative to our batch learning framework can be to mine all the
frequent patterns and select a relatively smaller subset of patterns for
getting a user's feedback and use these for learning in one iteration. This
approach is not scalable as the number of frequent patterns in ${\cal O}$ is
typically very large even for a moderate-size dataset. Besides, if the learning
entity receives the frequent patterns as a data stream the one-iteration
learning becomes infeasible, but a batch learning framework still works.

The IPIPD framework of \alg\ is partitioned into two main blocks: {\em
Learning} and {\em Discovery}~(Figure \ref{fig:framework}). {\em
Learning} learns a model for a user; it contains following five modules:
Frequent Pattern Miner (FPM), Learner, Feature Generator~(FG) and
Feedback-Collector~(FC), and Model. The FPM works as a bridge between the Learner and the
data. FPM module can be any off-the-shelf pattern mining algorithm that mines
frequent patterns given a normalized minimum support threshold. Any of the
existing off the shelf pattern mining algorithms ranging from
sequential~\cite{mybib:APRIORI} to randomized~\cite{Hasan.Zaki:09}, can be
used.  The Learner learns $u$'s interestingness function ($f$) over the
pattern-set ${\cal O}$ in multiple stages while utilizing user's feedback on a
small number of patterns selected from a partition of ${\cal O}$ in each stage.
The Feature Generator~(FG) module is responsible for generating
efficient and appropriate feature vector representation of the incoming
patterns for the Feedback Controller and the Model to use. 
FG uses the bag of words method for set patterns and NN-based unsupervised language
model for sequence and graph patterns. The Feedback-Collector~(FC)'s
responsibility is to identify the patterns using a exploitation-exploitation 
based strategy (explained in Section~\ref{sec:representative}), that are sent to the
user for feedback. Note that, patterns are sent to the user for feedback in its
original form, not in the feature vector representation.  {\em Discovery} delivers
personalized interesting patterns to the user using the learned model.

We like to highlight one relevant fact about the learning task of the proposed
framework. In the learning phase, the interaction between the FPM and the
Learner is one-way, from the FPM to the Learner, and the Learner has no
influence over FPM's mining process. This setup works well for large datasets
for which the mining task may take longer time, yet the learning task can start as
soon as the first batch of patterns is available.  Moreover, if a
user wants to end the learning session before FPM finishes, the learner can
still recommend patterns to the user based on the current learning model.  

\begin{figure}[!ht]
  \small
  \fbox{
  \begin{minipage}{0.45\textwidth}
      \begin{flushleft}
        {\bf PRIIME\_Learning}$({\cal D})$:\\
        {\bf comment:} ${\cal D}$ is the transaction dataset, \\
        1.~~$M_{feat}$ = {\bf Train feature generation model}(${\cal D}$)\\
        2.~~Initialize model parameters $\mathbf{w}$\\
        3.~~{\bf while} {1}:\\
        4.~~~~~~${\cal B}$ = Incoming batch of patterns from $FPM$\\
        5.~~~~~~$P_{feat}$ = {\bf Get feature representation}(${\cal B}$,$M_{feat}$)\\
        6.~~~~~~${\cal P}$ = {\bf Collect feedback}($P_{feat}$)\\
        7.~~~~~~$\mathbf{w}'$ = {\bf Train learning model}(${\cal P}$,$\mathbf{w}$)\\
        8.~~~~~~{\bf if} $||\mathbf{w}'$-$\mathbf{w}$|| < $threshold$:\\
        9.~~~~~~~~~~~{\bf break}\\
        10.~~~~~$\mathbf{w}$ = $\mathbf{w}'$\\
        11.~~{\bf return} $\mathbf{w}$\\
      \end{flushleft}     
  \end{minipage}
  }
\vspace{-0.05in}
  \caption{Pseudo code of Iterative Learning of \alg\ }
  \label{fig:alg}
\vspace{-0.15in}
\end{figure}

\section{Learning Method}\label{sec:method}

In \alg\ , the learning of interestingness function $f$ and pattern
mining by FPM proceeds in parallel over a few iterations. In the $i$'th
iteration, PM generates a partition (${\cal B}_i$) of frequent pattern set
${\cal O}$, and Learner selects $l$ patterns $\{p_t\}_{1\le t \le l}$ from
${\cal B}_i$ and send them to $u$ for rating.  Meanwhile, FG obtains
$\{\mathbf{x}_t\}_{1\le t \le l}$, the feature representation of the patterns 
$\{p_t\}_{1\le t \le l}$. Partitioning of frequent pattern
enables model learning to proceed even if all the frequent patterns are not
available, which is better than earlier works~\cite{Xin.Shen.ea:06} where model learning is
performed after the entire ${\cal O}$ is available. Learner collects
$u$'s rating $\{y_t=f(p_t)\}_{1\le t \le l}$ using an
interactive console. Then the Learner updates its
current model $f$ using $\langle \mathbf{x}_t, y_t \rangle$. 

In Figure~\ref{fig:alg}, we present the pseudo code
of \alg's learning phase. \alg\ starts by training the
unsupervised feature generation model $M_{feat}$ using the transaction data ${\cal D}$
and initializes the learning model $\mathbf{w}$. Afterward in each iteration, it uses
$M_{feat}$ to infer the feature representation of each pattern in the incoming
batch $\cal B$. Then the algorithm selects patterns for feedback and sends them to the user. 
Once it receives
the feedback, it updates the current learning model. The algorithm continues
until the improvement of the learning model falls below a user-defined threshold.
In the following, we discuss each of the tasks.

\subsection{Pattern Representation} \label{sec:data-rep}

Say ${\cal{D}} = (T_1,T_2,\cdots, T_N)$ is a transaction dataset.  ${\cal I}$
is the set of all single length frequent patterns.  ${\cal I}$ is a set of
item, a set of events, or a set of edges depending on whether ${\cal D}$ is an
itemset, a sequence or a graph dataset, respectively. For a given support
threshold, $\cal{O}$ is the set of all frequent patterns. For using a machine
learning method for finding interesting patterns, \alg\ first finds a feature
vector representation of these patterns.

\subsubsection{Set Patterns}
For a set data, each transaction $T_i$ is a collection of non-repetitive
items. Considering the dataset $\cal D$ as a text corpus and each transaction
in the corpus as a document, the set ${\cal I}$ can be represented as the dictionary
of frequent words i.e.~items. Following the bag of words model,
we represent each pattern $p \in {\cal O}$ as a binary
vector $V$ of size $|{\cal I }|$, constructed as below: 
\[
\small V(i)= 
\begin{cases} 
1, & \text{if item $i$} \in {p}\\ 
0, & \text{otherwise} \\ 
\end{cases} 
\] 

Such a representation maps each pattern to a data point $\mathbf{x} \in
{\{0,1\}}^{|{\cal I}|}$ space.  For example, consider a dataset in which $A$,
$B$, $C$, and $E$ are frequent items. Thus ${\cal I} = \{A, B, C, E\}$; A
frequent pattern $AB$ will be represented by the vector $(1, 1, 0, 0)$; a
pattern $ACE$ will be represented by the vector $(1, 0, 1, 1)$ and so on.
 
\begin{figure}[!ht]
\centering 
\includegraphics[width=0.5\textwidth]{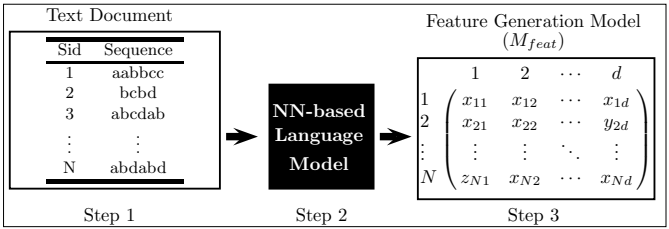}
\vspace{-0.15in} 
\caption{Unsupervised Feature Construction of Sequence
Patterns} 
\vspace{-0.1in} 
\label{fig:seq_rep} 
\end{figure}

\subsubsection{Sequence Pattern}

In sequence dataset $\cal D$, each transaction $T_i$ is an ordered list of events.
For example, $ACCGA$ is sequence and $A$, $C$ and $G$ are events. Similar to
set, One can view the dataset $\cal D$ as a text corpus and sequence of events
in a transaction as a sentence in a language, where the events are the words.
Given a set of sentences, our objective is to find a language model
in which a sequence has a metric embedding in an appropriately
chosen vector space. To obtain the language model, we apply \texttt{Paragraph Vector}~\cite{le2014distributed}. 
It finds the $d$-dimensional ($d$ is user defined) feature representation of all sequences in
$\cal D$.

In Figure~\ref{fig:seq_rep}, we illustrate how we compute feature
representation of the sequences in $\cal D$.
As we can see, we treat sequences
as sentences in a text corpus~(step 1 in Figure~\ref{fig:seq_rep}). In the 2nd
step, we pass the corpus in the language model~(shown as a black box in
Figure~\ref{fig:seq_rep}). As an output~(step 3 in Figure~\ref{fig:seq_rep}), 
language model produces feature
representation of a given length~($d-$dimension) for all transactions, i.e.
sequences~($1,2,\cdots N$). Note that, training of such language model
is computationally expensive and perform training in each iteration of \alg\
with the incoming batch of patterns is inefficient. As an alternative, \alg\ trains 
the model say, $M_{feat}$ using the dataset $\cal D$
as training data~(Line 1 in Figure~\ref{fig:alg}). Afterward in each iteration, 
it populates $d$-dimensional feature vector of incoming patterns by using $M_{feat}$.

 \begin{figure}[!ht]
\centering
\includegraphics[width=0.5\textwidth]{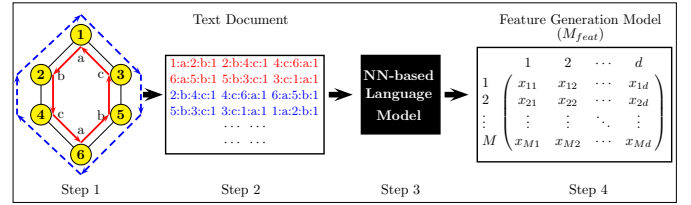}
\vspace{-0.15in}
\caption{Unsupervised Feature Construction of Graph Patterns}
\vspace{-0.1in}
\label{fig:graph_rep}
\end{figure}

\subsubsection{Graph Pattern}

In graph dataset $\cal D$, each transaction $T_i$ is a labeled, undirected and
connected graph.  In order to use neural network based unsupervised language
model, we have to map the dataset $\cal D$ to a text corpus and the graphs in
${\cal D}$ to sentences, on which we apply \texttt{Paragraph Vector}~\cite{le2014distributed}.
For converting a graph into sentence(s)  we break the graph into a collection
of simple paths, such that edge ordering is maintained in each path. Each path
can then be viewed as a sentence and the edges in the path as words. To
construct these paths from a graph $T_i$, for each node $u$ in $T_i$, we
perform a dfs-walk originated from $u$ that traverse all the edges in $T_i$.
The motivation of using dfs-walk is that it preserves topological information
of a graph around the originated vertex; multiple dfs-walks originated from
different vertices capture topological information of the entire graph.
Besides, it provides an ordering of edges in the paths which serves as
sentences. The number of paths that we generate from each graph is exactly equal
to the number of vertices in the graph because each dfs-path is originated
from a distinct vertex in the graph. 

Note that, for a set or a sequence dataset, we maintain a one-to-one mapping
between the transactions and the sentences, but for graph, it is a one-to-many
mapping. The primary reason for this is that a
complex object like graphs cannot be represented by only one sequence of
vertices. More importantly, the number of transactions for graph data is
usually small~(Table~\ref{tab:dataset}) compared to a dataset of sets and
sequences.  So populating one sentence per graph is not an optimal choice.  In
many of the existing works on machine learning, specifically in deep learning,
artificial random noises are inserted to create distorted
samples~\cite{Simard.Steinkraus:2003,ciresan:2012} which greatly increases the
training set size. In~\cite{Simard.Steinkraus:2003}, authors proposed a
technique called ``elastic distortion'' to increase the number of image
training data by applying simple distortions such as translations, rotations,
and skewing.  However in the existing works, no such mechanism is available
for training data inflation for a graph data. The one-to-many mapping that
we use is such a mechanism which enables us to create many samples for a 
graph instance, thus increasing
the learning ability of the model for graph patterns.

In Figure~\ref{fig:graph_rep}, we illustrate the entire process. In Step 1, we
have a graph~(transaction) with $6$ nodes and $6$ edges, where each of the
nodes has a label. We assume all the edges share the same label $1$ (not shown
in the figure). Step 2 obtains six dfs-walk edge sequences, on which two are
shown in the figure---one dfs-walk originated from  vertex $1$~(red ink) and
the other originated from vertex $2$~(blue ink and dashed).  Note that, it is possible to
have dfs-walk with different ordering from a vertex but in this work, we allow
one walk per vertex. Once we have such representation of the entire data $\cal
D$, we feed the text corpus to the language model~(Step 3) and finally for a
given feature length $d$, it trains the model~($M_{feat}$) and produces feature
vectors of all $M$ dfs-walks in the text corpus.  During the classification stage
for a pattern graph, we create only one dfs-walk from a randomly selected
vertex and generate feature vector representation by using $M_{feat}$.

We give some perspective of adapting neural network based unsupervised 
language modeler~\cite{le2014distributed} to model and find feature representation of a
graph. Assume a sequence of words $W = \{w_0,w_1,\cdots,w_n\}$, where $w_i \in
V$ ($V$ is the vocabulary) and a paragraph $P$ containing the sequence of
words, a language model maximizes $\text{Pr}[w_n | w_0,w_1,\cdots,w_{n-1},P]$
over all the training corpus. $P$ can be thought of as another word, which
behaves as a memory that recalls what is missing from the current context~($
w_0,w_1,\cdots,w_{n-1}$). When we map dfs-walks as sentences, estimated
likelihood can be written as $\text{Pr}[v_i | v_0,v_1,\cdots,v_{i-1},
d_{walk}]$, which is the likelihood of observing a vertex $v_i$ in a walk given
all the previously visited vertices and the walk $d_{walk}$. Note that, $v_i$
in the likelihood function does not necessary have to be at the end of the
context~($v_0,v_1,\cdots,v_{i-1}$), rather a context of a vertex consists of
vertices appearing to the left and right of the given vertex in the dfs-walk.
The training of word vectors and paragraph vectors is done using stochastic
gradient descent and the gradient is obtained via back-propagation. Please
see~\cite{le2014distributed} for more detail of the method.

\subsection{Classification Model}\label{sec:model}

The \alg\ models the user's interestingness profile over the patterns
as a classification problem and learns the model parameters through an interactive feedback
mechanism. 
In each iteration, \alg\ executes a supervised training session of
the classification model using the corresponding data points of the released
patterns to the user. We use multinomial logistic regression
as the classification model $H_{\boldsymbol{\theta}}$, which is known as
softmax regression in the literature.  
In multinomial logistic regression, the
probability that a data point $\mathbf{x}_i \in \mathbb{R}^d$ belongs to class $j$ can
be written as, 
\begin{equation}
\small
 H_{\boldsymbol{\theta}}(\mathbf{x}_i) = p(y_i =
j|\mathbf{x}_i;\boldsymbol{\theta}) =
\frac{exp({\boldsymbol{\theta}_{j}}^T\mathbf{x}_i)}{\sum_{l=1}^c
exp({\boldsymbol{\theta}_{l}}^T\mathbf{x}_i)}
\end{equation}
 where, $j \in \{1,\cdots,c\}$ is the set
of class labels and $\boldsymbol{\theta}_j$ is the weight vector corresponding to
class $j$. The prediction of the model can be computed as, $\hat{y}_{i}=
{\argmax}_j~p (y_i = j|\mathbf{x}_i,\boldsymbol{\theta})$. Given a set of labeled
training data $X = \{(\mathbf{x}_1,y_1), . ., (\mathbf{x}_m, y_m)\}$, the weight matrix
$\boldsymbol{\theta} \in c \times d$ ($c$ is the number of class and $d$ is the
size of a feature vector) is computed by solving the convex optimization
problem as shown in Equation~\ref{eq:solve} using gradient descent. Note that,
each data points $\mathbf{x}_i \in \mathbb{R}^d$ in the training set corresponds to a
pattern.

\begin{multline}\label{eq:solve}
 \argmin_{\boldsymbol\theta} J(\boldsymbol\theta) = -\frac{1}{m} \sum_{i=1}^m \sum_{j=1}^c {\bf 1}_{\{y_i=j\}} \cdot \log~p(y_i = j|\mathbf{x}_i;\boldsymbol\theta) \\
 +\frac{\lambda}{2}\sum_{j=1}^{c}\sum_{k=1}^d\theta_{jk}
\end{multline}

Here $J(\boldsymbol\theta)$ is the notation to represent the cost function in Equation~\ref{eq:solve}, 
${\bf 1}_{\{y_i=j\}}$ is the indicator function indicating that only the output of the classifier 
corresponding to the correct class label is included in the cost.
$\frac{\lambda}{2}\sum_{j=1}^{c}\sum_{k=1}^d\theta_{jk}$ is the regularization term to avoid over-fitting.
In each iteration, \alg\ applies L-BFGS-B optimization function implemented
in python's scipy package to train the model.  

\subsection{Selection of Representative Data-points for Feedback}\label{sec:representative}

As discussed in Section~\ref{sec:introduction}, one of the reasons of the superior performance
of our proposed method over the existing ones is due to the exploitation-exploration
 based feedback collection. In each iteration,
for an incoming batch, \alg\ exploits the currently learned model~($M_{cur}$) 
to identify the patterns that will have the largest
impact on the parameters of the learning model. We can identify those patterns by checking 
whether these patterns make the maximum change in
the objective function. Since we train softmax classifier using gradient descent, we include
the patterns to the training set if it creates the greatest change in the
gradient of the objective function,
\begin{equation}
\bigtriangledown_{\theta_j}J(\boldsymbol\theta) = -\frac{1}{m} 
\sum_{i=1}^m \left[ \mathbf{x}_i\left({\bf 1}_{\{y_i=j\}} - p(y_i = j|\mathbf{x}_i;\boldsymbol\theta) \right) \right] + \lambda\theta_j 
\end{equation}
However, gradient change computation requires the knowledge of the label, which
we don't have. So, we compute the expected gradient length~\cite{Settles.Craven:2008} 
of $\mathbf{x}_i$ as shown in Equation~\ref{EGL},
\begin{equation}\label{EGL}
EGL(\mathbf{x}_i) = \sum_{j=1}^c p(y_i = j|\mathbf{x}_i;\boldsymbol\theta)||\bigtriangledown_{\theta_j}J(\boldsymbol\theta)||
\end{equation}
\textcolor{black}{
After exploitation step, \alg\ retains the top $50\%$ of patterns~($\mathbf{x}_i s$) according 
to the expected gradient length~(EGL) as candidates for feedback. 
However, Expected gradient length based approach enables \alg\ to identify patterns 
that can make greatest changes in the 
learning model, but it does not ensure diversity among the patterns in terms of its composition,
i.e. two patterns can be made of similar items, events or edges.
To ensures that a broader span of the pattern
space is covered, \alg\ uses $k$-center search~(exploration) to identify $k$ patterns 
for feedback from the candidates set.}

{\bf $k$-Center.} $k$-center method finds $k$ data points as centers so that
the maximum distance from each data point to its nearest center is minimized.
This $k$-center problem is NP-complete, but there exists a fast 2-approximation
greedy algorithm that solves this problem in $O(kn)$ time~\cite{Gonzalez:85}.
We set $k$ to 10 in all our experiments. For choosing $10$ representative patterns, 
we use the greedy $10$-center
algorithm considering the Jaccard distance for set patterns and euclidean distance 
for sequence and graph patterns. 

\subsection{Stopping Criteria}\label{sec:stop}
The stopping criterion is an important element of our proposed system. The execution of \alg\ 
can be halted from the system side as well as the user side. \alg\ sets a minimum default 
iteration counts to $10$ and after 10th iteration, it keeps track whether getting 
additional feedback impact significantly
on the learning model or not. If the improvement falls below to a threshold~($1E-4$),
the execution of the feedback session halts. 

\section{Experiments \& Results}\label{sec:experiment}

\subsection{Data}

We use four set datasets, two sequence datasets, and two graph datasets to
validate the performance of \alg. Two of the set datasets, \texttt{Chess} and
\texttt{Mushroom}, are real-life datasets collected from FIMI
repository\footnote{\url{http://fimi.ua.ac.be/data/}}. The third one is
an artificially generated electronic health record~(EHR) dataset created by Uri
Kartoun from Harvard University\footnote{\url{http://bit.ly/1Db1yHo}}, which
has health record of $10,000$ patients. To generate a class label for each
patient, we identify whether the patient is diagnosed with arthritis, type-I/II
diabetics or not~\footnote{Generally class label is not needed for pattern
mining. We use the class labels to simulate user's interestingness over the
patterns.}.  The last set dataset is a drug side effects dataset collected from
CIDER repository\footnote{\url{http://sideeffects.embl.de/download/}}. This
data contains $996$ drugs and associated side effects. We consider each drug as
a transaction and side effects as the items. Each drug also has a class label
based on the primary disease which it cures: cancer, heart diseases, brain diseases, 
pain, and others. 

The two sequence dataset
Reuters1 and Reuters2 is collected according to the procedure mention in a
sequence classification work~\cite{zhou2013itemset}. These datasets are prepared by
extracting specific keywords from the entries in the Reuters-21578 
corpus~\footnote{\url{http://web.ist.utl.pt/~acardoso/datasets/r8-train-stemmed.txt}}.
Reuters1 consists of word sequences labeled by either ``earn'' or ``acq''. Reuters2 
consists of sequences labeled by ``interest'', ``money-fx'', ``crude'' and ``trade''.
For more information see \cite{zhou2013itemset}. Out of the two graph datasets, one is
called \texttt{pdb}, which is collected from predictive toxicology 
challenge\footnote{\url{http://www.predictive-toxicology.org/ptc/}}. The other graph dataset, 
\texttt{Mutagenicity-II} is obtained from the authors of~\cite{mybib:MUTAGENI}.
In Table~\ref{tab:dataset}, we present some basic statistics about these four
datasets; $\rho^{\text{min}}$ is the minimum support threshold, and ${\cal C}$
is the set of all closed frequent patterns. For each dataset, we choose $\rho^{\text{min}}$
in a way so that we can have a large number of closed frequent patterns
for the validation of the system.

\begin{table}
\small
 \center
\begin{tabular}{ l l l l l l }\hlinewd{2pt}
			&			& \# of 	& Avg.			&  			&  		\\
Type			& Dataset		& trans-	& transac-  		& $\rho^{\textbf{min}}$ & $|{\cal C}|$ 	\\
			&			& actions	& tion size 		&  			& 		\\\hlinewd{1pt}
			& Chess 		& 3196 		& 37 			& 2100 			& 99262 	\\
Set			& Mushroom		& 8124 		& 22  			& 1100 			& 53628 	\\
			& EHR			& 10000 	& 43 			& 3000 			& 104297 	\\
			& Drug-Side-effects	& 996		& 99			& 220			& 117544	\\\hlinewd{0.5pt}
Sequence		& Reuters1		& 4335		& 56			& 86			& 54193		\\
			& Reuters2		& 900		& 108			& 36			& 48209		\\\hlinewd{0.5pt}
Graph			& PDB			& 416		& 15			& 8			& 9246		\\
			& Mutagenicity-II	& 684	 	& 24			& 8			& 20771		\\\hlinewd{2pt}
\end{tabular}
\vspace{-0.1in}
\caption{Dataset Statistics}
\label{tab:dataset}
\vspace{-0.1in}
\end{table}

\subsection{Experimental Setup}\label{sec:ex-setup}

An essential part of \alg\ is a pattern mining algorithm that mines frequent
patterns and send them to the Learner in batches. We use
LCM~\cite{Uno.Kiyomi:04}, spmf~\cite{spmf} and CloSpan~\cite{yan2003clospan}
for mining closed set, sequence and graph patterns, receptively. A closed
frequent pattern is defined as a pattern that does not have a
super-pattern~(super-set) that is frequent with the same support
count~\cite{zaki2014dataminingbook}. The set of closed pattern, ${\cal C}$, is
a subset of the set of frequent patterns, ${\cal O}$. Redundancy among the
patterns in ${\cal O}$ is substantially reduced in the patterns in ${\cal C}$,
so learning a function over the patterns in ${\cal C}$ is better than learning
a function over the patterns in ${\cal O}$.

For performance evaluation, we make 5-folds of the data of which 4 folds are
used for interactive training and 1 fold for testing. We use a grid search to
tune regularization parameter $\lambda$ and set it to $1$. We run
Gensim's~\cite{rehurek_lrec} implementation of \texttt{Paragraph Vector} model with the default
parameter settings, except for the parameter $d$ (the dimension of feature
vector), which we set it to $100$ for both sequence and graph after tuning it
using grid search. We report the performance result using average weighted F-score~(calculate
F-Score for each label, and compute their average, weighted by the number of
true instances for each label) over the folds. In each iteration, we
approximately use $2\%$ of the training data to construct the batch and select
$10$ patterns from the batch for feedback. Note that in the first iteration, \alg\ uses
the $k$-center search to find $10$ patterns for feedback. We implement \alg\
in python and run all experiments in 3 GHz Intel machine with 8GB
memory.

\subsection{Interestingness Criteria} 
Following earlier works~\cite{Xin.Shen.ea:06,Bhuiyan.Hasan:12}, for validating the 
proposed system we simulate the interestingness function of a user. 
The interestingness criteria of a pattern is derived from the 
class label of its predominating transactions. For example, if a pattern $p$ belongs 
to $t$ number of transactions among which $t_1$ and $t_2$ are the
numbers of transactions with label $1$ and $2$ respectively; then $t = t_1+t_2$. 
The corresponding label/feedback of $p$ is measured as $1$ if  $max(t_1,t_2)=t_1$
else $2$. 

\begin{figure}[!ht]
\vspace{-0.55in}
\centering
\includegraphics[width=0.45\textwidth]{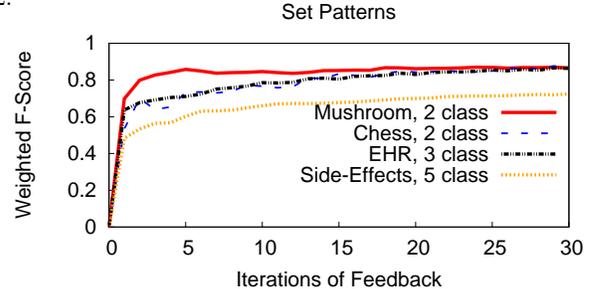}
\vspace{-0.4in}
\caption{Weighted F-Score of the learner across iterations of feedback in set dataset}
\vspace{-0.1in}
\label{fig:performance_set}
\end{figure}

\begin{figure}[!ht]
\vspace{-0.55in}
\centering
\includegraphics[width=0.45\textwidth]{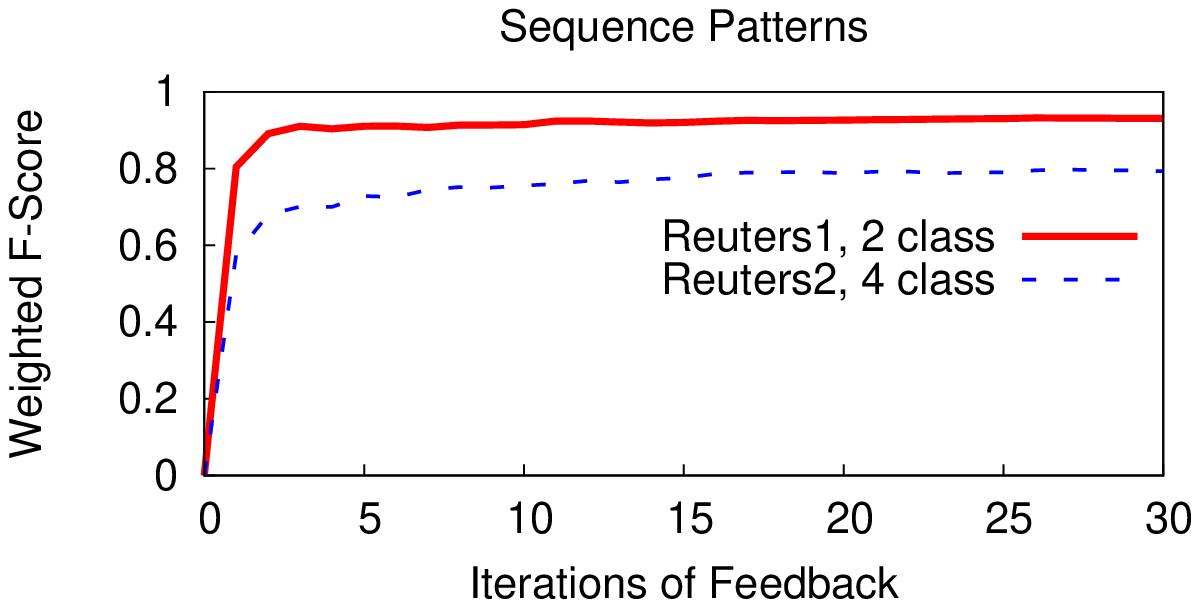}
\vspace{-0.4in}
\caption{Weighted F-Score of the learner across iterations of feedback in sequence dataset}
\vspace{-0.1in}
\label{fig:performance_seq}
\end{figure}

\begin{figure}[!ht]
\vspace{-0.40in}
\centering
\includegraphics[width=0.45\textwidth]{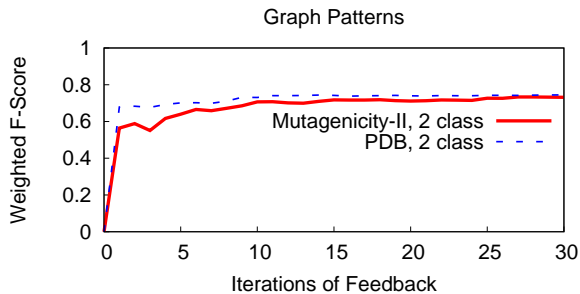}
\vspace{-0.4in}
\caption{Weighted F-Score of the learner across iterations of feedback in graph dataset}
\vspace{-0.1in}
\label{fig:performance_graph}
\end{figure}

\subsection{Experiment on the Learner's Performance}\label{sec:performance}
In this experiment, we evaluate the performance of softmax 
based learning algorithm using weighted F-Score metric.
we show how the performance of the learning model improves with the iterations of feedback. 
As discussed earlier in each iteration, the learner selects 
ten patterns for feedback and iteratively updates the current model.
After each iteration, we measure weighted F-Score of our currently learned model with the test
fold. Then we plot F-score of the learner across the number of iterations
executed so far. In Figure~\ref{fig:performance_set},~\ref{fig:performance_seq} and \ref{fig:performance_graph},
we show the progression of F-score across iterations for the set, sequence and
graph datasets. The performance on set and sequence datasets is relatively better than
that on graph datasets. The reason behind this is that effective feature vector
construction for graph patterns is comparatively difficult than set and sequence patterns.
For all the datasets, within five to ten iterations of feedback (each iteration has 10 patterns),
learning method converges and stay stable. Compared to the entire (closed) frequent pattern
space, the number of feedback that \alg\ use is almost negligible, yet the performance of
the learning model is satisfactory.

\begin{table}[!h]
\vspace{-0.1in}
 \center
 \small
\begin{tabular}{l l l l }\hlinewd{2pt}
Dataset			& Accuracy       		& Accuracy~			& Accuracy\\
			& \alg 's			&(\cite{Xin.Shen.ea:06}'s)	&\cite{Bhuiyan.Hasan:12}'s\\\hlinewd{1pt}
Chess			& 90.8 \% 			& 42.7\%			& 55.5\%\\
Mushroom		& 84.5\% 			& 45.3\% 			& 61.4\%\\
EHR			& 78.8\% 			& 46.6\% 			& 40.6\%\\
Drug-Side-Effects	& 71.9\% 			& 36.3\% 			& 54.50\%\\\hlinewd{0.5pt}
Reuters1		& 91.1\%			& 35.9\%			& NA		\\
Reuters2		& 76.6\%			& 38.2\%			& NA		\\\hlinewd{0.5pt}
PDB			& 73.1\%			& NA				& 50.7\%		\\
Mutagenicity-II		& 66.8\%			& NA				& 44.7\%		\\\hlinewd{1pt}
\end{tabular}
\caption{Comparison on percentage accuracy of our algorithm with the existing ones } 
  \vspace{-0.17in}
\label{tab:compare}
\end{table}

\subsection{Comparison with the Existing Algorithms}\label{sec:compare}
To compare with the existing works we implement \cite{Xin.Shen.ea:06}'s
interactive learning method in python and compute the percentage accuracy
experiment of their method for the set and sequence data by following the instructions 
provided in the paper~\cite{Xin.Shen.ea:06}.  We also compare with \cite{Bhuiyan.Hasan:12}'s
sampling algorithm~(executable collected from the authors)
for the set and graph data; we pick sampler's recall metric mentioned in
\cite{Bhuiyan.Hasan:12} as it computes the percentage of interesting patterns
recovered by the sampler. For all algorithms, we use 100 feedback and use the
percentage of accuracy as a comparative metric.  Table~\ref{tab:compare} shows the comparison, which
clearly demonstrates that our algorithm performs substantially better than the
existing methods. 
We validate that this performance boost can be credited to efficient feature
construction for complex patterns and intelligent exploitation-exploration strategy
for feedback pattern selection. We will discuss more on this in the following sections. 

\begin{figure}[!ht]
\vspace{-0.1in}
\centering
\includegraphics[width=0.45\textwidth]{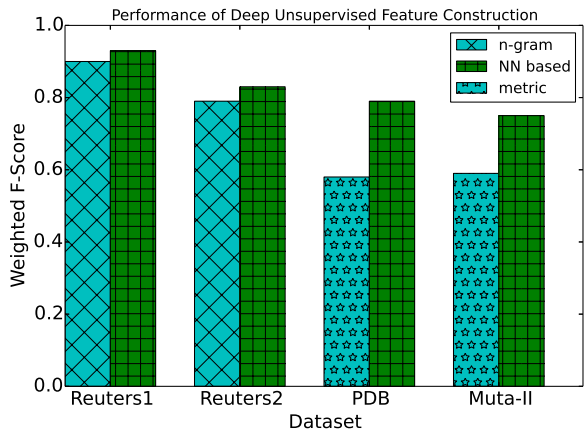}
\vspace{-0.2in}
\caption{Performance of Deep Unsupervised feature construction. }
\vspace{-0.15in}
\label{fig:doc2vec_performance}
\end{figure}

\subsection{Comparison of Unsupervised Feature Learning Approach with
Others}\label{sec:feature_cons} 

In this experiment, we evaluate the effectiveness of the proposed unsupervised
feature construction based approach with alternative feature representation for
sequence and graph patterns. For sequence pattern, we use n-gram based
technique, where we extract all 2-grams and 3-grams from the data and use those
as features.  For graph patterns, we obtain $20$ topological metrics compiled
by~\cite{Li:2012} and consider these metrics as features for graph patterns.
We run \alg\ for $10$ iterative sessions~(100 feedback) and use the weighted
F-score for comparison.  In figure~\ref{fig:doc2vec_performance}, we show the
findings using bar chart. First two groups of the bar chart are for sequence datasets and
the remaining two groups are for graph datasets.
As we can see,  \alg's unsupervised feature learning performs 
better than the alternative feature representation of both sequence and graph patterns. 
Specifically,  \alg's feature embedding for graph datasets is significantly 
better than that of the competitors'.

\begin{figure}[!ht]
\vspace{-0.45in}
\centering
\includegraphics[width=0.45\textwidth]{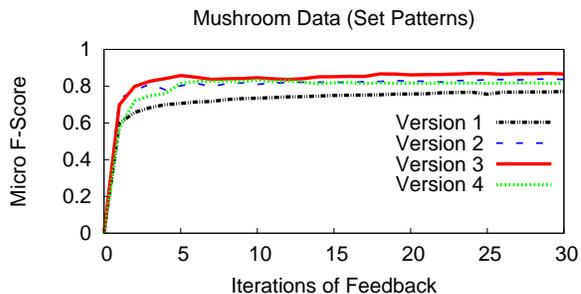}
\vspace{-0.4in}
\caption{Performance of the learner with different feedback collection scheme in Mushroom set dataset}
\vspace{-0.1in}
\label{fig:performance_feedbackset}
\end{figure}

\begin{figure}[!ht]
\vspace{-0.45in}
\centering
\includegraphics[width=0.45\textwidth]{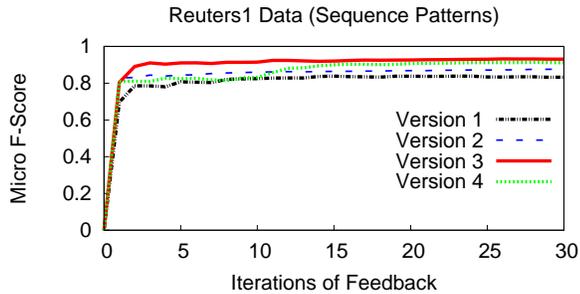}
\vspace{-0.4in}
\caption{Performance of the learner with different feedback collection scheme in Reuters1 sequence dataset}
\vspace{-0.1in}
\label{fig:performance_feedbackseq}
\end{figure}

\begin{figure}[!ht]
\vspace{-0.40in}
\centering
\includegraphics[width=0.45\textwidth]{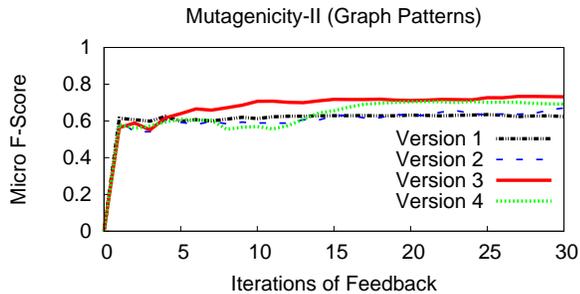}
\vspace{-0.4in}
\caption{Performance of the learner with different feedback collection scheme in Mutagenicity-II graph dataset}
\vspace{-0.1in}
\label{fig:performance_feedbackgraph}
\end{figure}

\subsection{Representative Patterns Selection} \label{sec:select}
As discussed in Section~\ref{sec:introduction}, our proposed solution is
different than the existing works on how we select patterns for feedback. 
To show that our approach is more effective, we create four different versions of
\alg\ . In one, we solely rely on exploitation
i.e pick top $10$ patterns for feedback from a batch according to the expected gradient length~(EGL). 
In second, we concentrate on exploration and select $10$ most diverse patterns for feedback
using only k-center from the entire batch. The third version is the one that we 
proposed in this work, where we retain top 50\% patterns from a batch according 
to the EGL then apply k-center to find $10$ most diverse patterns for feedback. 
In fourth, we use only k-center~(exploration) in the first $10$ interactive sessions, 
then we use the combination of EGL and k-center~(exploitation-exploration)
like third version for the next 10 iterations, and in final 10 iterations, 
we only use EGL~(exploitation).
In Figure~\ref{fig:performance_feedbackset},~\ref{fig:performance_feedbackseq}
and \ref{fig:performance_feedbackgraph}, we show the performance of the above four versions
of \alg\ for a set, sequence, and graph data. As we can see, for all the cases, our proposed
exploitation-exploration~(Version 3 curve in the figures) strategy performs better than the 
other three options. Our's performance better than the fourth version (Version 4 in the figures) 
establish the fact that in interactive pattern discovery, balancing exploration and exploitation
in all iterations of learning is beneficial than gradually changing from exploration
to exploitation like the fourth version.
More importantly, the performance of the learning methods that solely depends on the current
learning~(Version 1 in the Figures) is the lowest among all three, which suggests that high exploitation
of the current model for selecting patterns for feedback may yield sub-optimal model.


\section{Case Study}\label{sec:casestudy}
To demonstrate the applicability of interactive personalized interesting pattern discovery
in a real world setting, we perform a case study in the real-estate domain.
Generally, the success of a home searching process depends on a user's ability to construct
a well-defined search query. It also depends on her patience while going through 
the hundreds of houses that are returned by the search engine. 
This assessment is particularly 
challenging for new home buyers, who are, sometimes not even familiar with all 
different home features. The overall process of selecting the right home takes time; actually, 
for 40\% of first time home
buyers, the lag time between research and action (buying a home) is around 120
days~\cite{Realtors:2013}. Given that home buying is a significant investment,
most home buyers take help from an experience real estate agent who can read
the buyer's mind as soon as possible and show her the home that is just right
for her.

\alg\ in this setup works just
like a virtual real estate agent for the home buyers; where
each house can be thought of as a set type transaction and house 
features as the items in the transaction. In each iteration,
\alg\ presents a set of frequent patterns which are
a summary set of features of the houses in the data. By utilizing
the feedback over the quality of patterns \alg\ gradually
learns the user's interestingness criteria on the house features.
Finally, using the learned model \alg\ identifies the house summaries~(patterns)
as well as the houses that the user will prefer. 
In the following, we will explain in detail from the data collection,
experimental setup, and observations.

\subsection{Data}
We crawl \texttt{trulia.com} from November 2015 to January 2016~(3 months) 
for five major cities: \texttt{Carmel}, \texttt{Fishers}, \texttt{Indianapolis}, \texttt{Zionsville} and 
\texttt{Noblesville} in the Central Indiana. The information that we crawl is the
basic house information, text on house detail, school, and crime information. 
In total, we crawl $7216$ houses among which $596$ are from Carmel,
$525$ are from Fishers, $5485$ are from Indianapolis, $211$ are from Zionsville and
$389$ are from Noblesville. We choose Central Indiana home market because of
our familiarity with the region.

To clean the crawled text data we use standardized data cleaning approach,
i.e. remove stop words, stemming and lemmatization of words. We have manually
cleaned some keywords that are written in unstructured abbreviated
form. For example, keyword \texttt{fireplace} is written as \texttt{frplc}, or \texttt{firplc}
in many house details.
In order to compile house features set, we use Rake~\cite{rose2010automatic} 
to find the key phrases. Next we group the key phrases, where we keep two key phrases together
if they share at least one keyword. That way, we get all the features of a category together.
For example, all the key phrases related to \texttt{basement} grouped together. Finally,
we manually investigate these groups and identify specific house features. In the end, we
retain only those house features that appear in at least ten~(10) houses.
The total number of house features we extracted from the crawled data is 130 that
includes interior features, neighborhood features, and school information.

\subsection{Setup}
We empirically evaluate \alg\ by demonstrating its performance over a
demographic group of home buyers. We identified a group from
the yearly report of \url{realtor.com}~\cite{Realtors:2015}
and populate a potential house features set that this group might like. The group is a
median-income couple with kids. People in this demographic group are of age between $35$ to $49$.
We select first $4$ features related to the type of a house and nearby schools quality. 
Next $3$ features related to the neighborhood of a house. Third set of features are related 
to the interior of a house. According to the report~\cite{Realtors:2015}, 
$85\%$ of the peoples in the group prefer single family home and 
$43\%$ of the individuals in this group want good quality nearby schools, 
so we select \texttt{Single Family Home}, \texttt{Nearby schools}, \texttt{Above Average Quality Primary School}, 
and \texttt{Above Average Quality Middle School} as house features. 
$24\%$ of families with kids prefer nearby parks, playground, so
we pick \texttt{Neighborhood playground}, \texttt{Trails}, \texttt{Community pool} and \texttt{Lowest crime rate}.
$50\%$ of this demography have kids and most likely they prefer 
houses with open concept design, spacious, easy to maintain, common and work area, 
and large kitchen with necessary amenities. With that in mind, we pick,
\texttt{Open concept}, \texttt{4 Bedrooms}, \texttt{3 Bathrooms}, \texttt{2 car garage}, \texttt{finished basement},
\texttt{Large Master Bedroom}, \texttt{Hardwood floor}, \texttt{Open Concept kitchen},\\
\texttt{Stainless-steel appliance}, \texttt{Granite countertop and island}, \texttt{Dual sink},
\texttt{Many cabinets}, \texttt{Common area}, \texttt{Work area} and \texttt{Large backyard}.

\begin{figure}[!ht]
\vspace{-0.45in}
\centering
\includegraphics[width=0.45\textwidth]{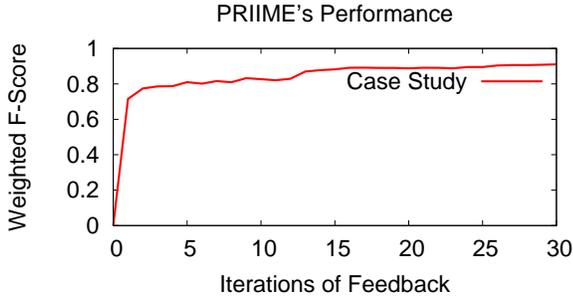}
\vspace{-0.35in}
\caption{Performance of the learner for the Demographic Group}
\vspace{-0.1in}
\label{fig:case_1_perform}
\end{figure}

\begin{figure}[!ht]
\vspace{-0.7in}
\centering
\includegraphics[width=0.45\textwidth]{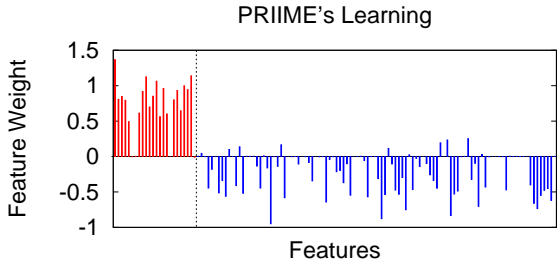}
\vspace{-0.45in}
\caption{Quality of Training on the Demographic Group}
\vspace{-0.1in}
\label{fig:case_1_learning}
\end{figure}


\subsection{Experiment and Observations}
After generating features set associated with the 
demographic group, we first create labels for the mined closed patterns from the 
data with $5\%$ support. We assign label 1~(interesting) to a pattern if it
contains $80\%$ or more features from the devised features set, otherwise 0~(non-interesting). 
Next, we split the patterns set into train and test and execute \alg\ 
over the training split. We set the batch size to $2000$ and 
consider $10$ patterns for feedback in each iteration. We measure the classification performance 
of \alg\ in terms of weighted F-Score. 

First, we observe how \alg's performance changes with the interactive sessions. 
In Figure~\ref{fig:case_1_perform}, we plot \alg's performance using 
weighted F-Score over the test data in the y-axis, 
and plot iterations count of \alg\ in the x-axis.
As we can see, the performance of the system improves with iterations of feedback.
After $5$ iterations, \alg\ reaches to a reasonable performance with a Weighted 
F-Score higher than $0.8$.
\alg\ reaches to a stable position in terms of performance after $10$ 
iterations.

To illustrate the training quality of the learning module in \alg, we create a bar
chart with the feature weights of the model. To visually show the distinction
between weights of the selected features set from the weight of the rest of the features, 
we divide the plotting area into two regions, in the left region we display the weights of the
selected features set in red ink and in the right region we show the 
weight of the rest of the features in blue ink. 
As we can see in Figure~\ref{fig:case_1_learning}, \alg\ puts positive
weights on the features from the selected set and negative weight to the majority of other features.
Few features in the right side of the plot have positive
weight and we found that top two~(according to weight)
of these features are \texttt{many windows} and \texttt{counter space} which
are related to the selected \texttt{open concept} and \texttt{open concept kitchen}
feature. Note that, the learned features weights used in the bar chart is collected after $10$ iterations. 


\section{Related Works}\label{sec:related}

There exist a few works which target personalized pattern discovery by using
users feedback.  In~\cite{Xin.Shen.ea:06} the authors propose a model that
learns a user's ranking function over the frequent itemsets.  They start with
the entire set of closed frequent itemsets, and send users a small list of
patterns; the user provides a complete ordering among the patterns which the
model uses as constraints in an SVM ranking based model. The model chooses the
next batch of patterns for feedback using the learned SVM ranking model.
Bhuiyan et al.~\cite{Bhuiyan.Hasan:12} propose another interactive pattern
mining framework, where the user interacts with a Markov Chain Monte Carlo
(MCMC) sampling entity, which samples frequent patterns from hidden data. The
target distribution of the sampler is iteratively refined based on binary
feedback from the user.  Boley et al. \cite{boley.Mampaey:2013} propose a
co-active based approach for interactive pattern mining, which use user's
feedback to decide between searching and sampling of patterns from the output
set.  There also exist a few other works~\cite{Mampaey.Tatti.ea:11}, which mine
a small set of interesting patterns by defining novel interestingness metrics
for frequent patterns. For instance, Mampaey et al.\cite{Mampaey.Tatti.ea:11}
summarizes the frequent pattern set and Blei et.  al.\cite{Bie:11} defines
subjective interestingness, both using maximum entropy model. \cite{Goethals.Moens.ea:11}
develops a toolbox with interestingness measures, mining and
post-processing algorithms as built-ins that can assist a user to visually
mine interesting patterns. However, the
scope of these works is different than our work.

There are some recent works on interactive knowledge discovery which solves
domain specific problems.  Examples include mining geospatial
redescriptions~\cite{Galbrun.Miettinen:2012}, and
subgroup
discovery~\cite{Dzyuba.Leeuwen.ea:2013,omidvar2015interactive}.
In the work on mining geospatial redescriptions
~\cite{Galbrun.Miettinen:2012}, the authors propose a
system called SIREN, in which a user builds queries based on his interests and
the system answers the queries through visualization of
redescriptions~(different ways of characterizing the same things).
In~\cite{Dzyuba.Leeuwen.ea:2013,omidvar2015interactive},
the authors present a framework which utilizes user's feedback and devises a
search procedure for finding subgroups.

\section{Conclusion}\label{sec:conclusion}

In this work, we propose a generic framework of interactive personalized interesting
pattern discovery called \alg . The proposed method 
uses user's rating on a small collection of
patterns for learning a user profile model, which at a later stage can be used
for recommending patterns that best align with user's interests.
Such a method is highly useful for identifying a small number of {\em
interesting} patterns where interestingness is defined through user's
rating.

\bibliographystyle{abbrv}
\balance
\bibliography{mybib}

\begin{thebibliography}{10}

\bibitem{Gonzalez:85}
Clustering to minimize the maximum intercluster distance.
\newblock {\em Theoretical Computer Science}, 38(0):293--306, 1985.

\bibitem{aggarwal2014data}
C.~C. Aggarwal.
\newblock {\em Data classification: algorithms and applications}.
\newblock CRC Press, 2014.

\bibitem{mybib:APRIORI}
R.~Agrawal and R.~Srikant.
\newblock Fast algorithms for mining association rules in large databases.
\newblock In {\em Proc. of VLDB}, pages {487--499}, 1994.

\bibitem{Bhuiyan.Hasan:12}
M.~Bhuiyan, S.~Mukhopadhyay, and M.~A. Hasan.
\newblock Interactive pattern mining on hidden data: A sampling-based solution.
\newblock In {\em Proceedings of the 21st ACM CIKM}, pages 95--104, 2012.

\bibitem{Bie:11}
T.~D. Bie.
\newblock Maximum entropy models and subjective interestingness: an application
  to tiles in binary databases.
\newblock {\em Data Mining and Knowledge Discovery}.

\bibitem{boley.Mampaey:2013}
M.~Boley, M.~Mampaey, B.~Kang, P.~Tokmakov, and S.~Wrobel.
\newblock One click mining-interactive local pattern discovery through implicit
  preference and performance learning.
\newblock In {\em KDD 2013 Workshop IDEA}, 2013.

\bibitem{Bondu.Lemaire.ea:10}
A.~Bondu, V.~Lemaire, and M.~Boulle.
\newblock Exploration vs. exploitation in active learning : A bayesian
  approach.
\newblock In {\em Int. Joint Conf. on Neural Networks}, pages 1--7, 2010.

\bibitem{mybib:MUTAGENI}
S.~Bringmann, A.~Zimmermann, L.~Raedt, and S.~Nijssen.
\newblock Don't be afraid of simpler pattern.
\newblock In {\em PKDD}, pages 55--66, 2004.

\bibitem{ciresan:2012}
D.~Ciresan, U.~Meier, and J.~Schmidhuber.
\newblock Multi-column deep neural networks for image classification.
\newblock In {\em Computer Vision and Pattern Recognition (CVPR), 2012 IEEE
  Conference on}, pages 3642--3649, 2012.

\bibitem{Dzyuba.Leeuwen.ea:2013}
V.~Dzyuba and M.~van Leeuwen.
\newblock Interactive discovery of interesting subgroup sets.
\newblock In {\em Advances in Intelligent Data Analysis XII}, volume 8207,
  pages 150--161. 2013.

\bibitem{spmf}
P.~Fournier-Viger, A.~Gomariz, T.~Gueniche, A.~Soltani, C.~Wu., and V.~S.
  Tseng.
\newblock {SPMF: a Java Open-Source Pattern Mining Library}.
\newblock {\em Journal of ML Research (JMLR)}, 15:3389--3393, 2014.

\bibitem{Realtors:2013}
A.~J.~S. from The National Association~of Realtors® and Google.
\newblock The digital house hunt: Consumer and market trends in real estate.
\newblock {\em National Association of Realtors}, 2013.

\bibitem{Galbrun.Miettinen:2012}
E.~Galbrun and P.~Miettinen.
\newblock Siren: An interactive tool for mining and visualizing geospatial
  redescriptions.
\newblock In {\em Proceedings of the 18th ACM SIGKDD}, pages 1544--1547, 2012.

\bibitem{Goethals.Moens.ea:11}
B.~Goethals, S.~Moens, and J.~Vreeken.
\newblock Mime: A framework for interactive visual pattern mining.
\newblock In {\em ECML}, pages 757 --760, 2011.

\bibitem{Hasan.Zaki:09}
M.~A. Hasan and M.~J. Zaki.
\newblock Output space sampling for graph patterns.
\newblock In {\em Proc. of VLDB}, pages 730--741, 2009.

\bibitem{le2014distributed}
Q.~V. Le and T.~Mikolov.
\newblock Distributed representations of sentences and documents.
\newblock {\em arXiv preprint arXiv:1405.4053}, 2014.

\bibitem{Li:2012}
G.~Li, M.~Semerci, B.~Yener, and M.~J. Zaki.
\newblock Effective graph classification based on topological and label
  attributes.
\newblock {\em Stat. Anal. Data Min.}, 5(4):265--283, 2012.

\bibitem{Mampaey.Tatti.ea:11}
M.~Mampaey, N.~Tatti, and J.~Vreeken.
\newblock Tell me what i need to know: succinctly summarizing data with
  itemsets.
\newblock In {\em Proc. of the 17th ACM SIGKDD}.

\bibitem{Mampaey.Vreeken:2012}
M.~Mampaey, J.~Vreeken, and N.~Tatti.
\newblock Summarizing data succinctly with the most informative itemsets.
\newblock {\em ACM Trans. Knowl. Discov. Data}, 6(4):16:1--16:42, 2012.

\bibitem{Mikolov:2013}
T.~Mikolov, I.~Sutskever, K.~Chen, G.~S. Corrado, and J.~Dean.
\newblock Distributed representations of words and phrases and their
  compositionality.
\newblock In {\em Advances in Neural Information Processing Systems 26}, pages
  3111--3119. 2013.

\bibitem{Realtors:2015}
T.~N.~A. of~Realtors.
\newblock 2015 home buyer and seller generational trends.

\bibitem{omidvar2015interactive}
B.~Omidvar-Tehrani, S.~Amer-Yahia, and A.~Termier.
\newblock Interactive user group analysis.
\newblock In {\em Proceedings of the 24th CIKM}, pages 403--412, 2015.

\bibitem{Perozzi:2014}
B.~Perozzi, R.~Al-Rfou, and S.~Skiena.
\newblock Deepwalk: Online learning of social representations.
\newblock In {\em Proceedings of the 20th ACM SIGKDD}.

\bibitem{rehurek_lrec}
R.~{\v R}eh{\r u}{\v r}ek and P.~Sojka.
\newblock {Software Framework for Topic Modelling with Large Corpora}.
\newblock In {\em {Proceedings of the LREC 2010 Workshop on New Challenges for
  NLP Frameworks}}, pages 45--50.

\bibitem{rose2010automatic}
S.~Rose, D.~Engel, N.~Cramer, and W.~Cowley.
\newblock Automatic keyword extraction from individual documents.
\newblock {\em Text Mining}, pages 1--20, 2010.

\bibitem{settles2010active}
B.~Settles.
\newblock Active learning literature survey.
\newblock {\em University of Wisconsin, Madison}, 52(55-66):11, 2010.

\bibitem{Settles.Craven:2008}
B.~Settles, M.~Craven, and S.~Ray.
\newblock Multiple-instance active learning.
\newblock In {\em Advances in Neural Information Processing Systems 20}, pages
  1289--1296. 2008.

\bibitem{Simard.Steinkraus:2003}
P.~Simard, D.~Steinkraus, and J.~C. Platt.
\newblock Best practices for convolutional neural networks applied to visual
  document analysis.
\newblock In {\em Document Analysis and Recognition, 2003. Proceedings. Seventh
  International Conference on}, pages 958--963, 2003.

\bibitem{socher:2011}
R.~Socher, E.~H. Huang, J.~Pennin, C.~D. Manning, and A.~Y. Ng.
\newblock Dynamic pooling and unfolding recursive autoencoders for paraphrase
  detection.
\newblock In {\em Advances in Neural Information Processing Systems}, pages
  801--809, 2011.

\bibitem{Szegedy:2014}
C.~Szegedy, W.~Liu, Y.~Jia, P.~Sermanet, S.~Reed, D.~Anguelov, D.~Erhan,
  V.~Vanhoucke, and A.~Rabinovich.
\newblock Going deeper with convolutions.
\newblock {\em CoRR}, abs/1409.4842, 2014.

\bibitem{Uno.Kiyomi:04}
T.~Uno, M.~Kiyomi, and H.~Arimura.
\newblock Lcm ver. 2: Efficient mining algorithms for frequent/closed/maximal
  itemsets.
\newblock In {\em Proc. IEEE ICDM'04 Workshop FIMI'04}, 2004.

\bibitem{Vreeken.Leeuwen:2011}
J.~Vreeken, M.~Leeuwen, and A.~Siebes.
\newblock Krimp: Mining itemsets that compress.
\newblock {\em Data Min. Knowl. Discov.}, 23(1):169--214, 2011.

\bibitem{Xin.Shen.ea:06}
D.~Xin, X.~Shen, Q.~Mei, and J.~Han.
\newblock Discovering interesting patterns through user's interactive feedback.
\newblock In {\em Proc. of the 12th ACM SIGKDD}, pages 773--778, 2006.

\bibitem{yan2003clospan}
X.~Yan, J.~Han, and R.~Afshar.
\newblock Clospan: Mining closed sequential patterns in large datasets.
\newblock In {\em In SDM}, pages 166--177. SIAM, 2003.

\bibitem{zaki2014dataminingbook}
M.~J. Zaki and J.~Wagner~Meira.
\newblock {\em Data Mining and Analysis: Fundamental Concepts and Algorithms}.
\newblock Cambridge University Press, 2014.

\bibitem{zhou2013itemset}
C.~Zhou, B.~Cule, and B.~Goethals.
\newblock Itemset based sequence classification.
\newblock In {\em Machine Learning and Knowledge Discovery in Databases}, pages
  353--368. Springer, 2013.

\end{thebibliography}
\end{document}